# MULTIMODAL RAG-DRIVEN ANOMALY DETECTION AND CLASSIFICATION IN LASER POWDER BED FUSION USING LARGE LANGUAGE MODELS

**Kiarash Naghavi Khanghah[1], Zhiling Chen[1], Lela Romeo[1], Qian Yang[2], Rajiv Malhotra[3], Farhad Imani[1], Hongyi Xu[1]***

[1] School of Mechanical, Aerospace and Manufacturing Engineering, University of Connecticut, Storrs, CT 06269
[2] School of Computing, University of Connecticut, Storrs, CT 06269
[3] Department of Mechanical & Aerospace Engineering, Rutgers, the State University of New Jersey, Piscataway, NJ 08854

* Email: hongyi.3.xu@uconn.edu

## ABSTRACT

*Additive manufacturing enables the fabrication of complex designs while minimizing waste, but faces challenges related to defects and process anomalies. This study presents a novel multimodal Retrieval-Augmented Generation-based framework that automates anomaly detection across various Additive Manufacturing processes leveraging retrieved information from literature, including images and descriptive text, rather than training datasets. This framework integrates text and image retrieval from scientific literature and multimodal generation models to perform zero-shot anomaly identification, classification, and explanation generation in a Laser Powder Bed Fusion setting. The proposed framework is evaluated on four L-PBF manufacturing datasets from Oak Ridge National Laboratory, featuring various printer makes, models, and materials. This evaluation demonstrates the framework's adaptability and generalizability across diverse images without requiring additional training. Comparative analysis using Qwen2-VL-2B and GPT-4o-mini as MLLM within the proposed framework highlights that GPT-4o-mini outperforms Qwen2-VL-2B and proportional random baseline in manufacturing anomalies classification. Additionally, the evaluation of the RAG system confirms that incorporating retrieval mechanisms improves average accuracy by 12% by reducing the risk of hallucination and providing additional information. The proposed framework can be continuously updated by integrating emerging research, allowing seamless adaptation to the evolving landscape of AM technologies. This scalable, automated, and zero-shot-capable framework streamlines AM anomaly analysis, enhancing efficiency and accuracy.*

**Keywords**: Multimodal Large Language Model; Laser Power Bed Fusion (L-PBF), Retrieval Augmented Generation, Additive Manufacturing

## 1. INTRODUCTION

The objective of this work is to evaluate the feasibility of establishing a generative model that leverages literature-based information to detect and classify anomalies in unseen material images, without relying on in-house experimental data. Specifically, we focus on anomalies in Additive manufacturing (AM) processes.

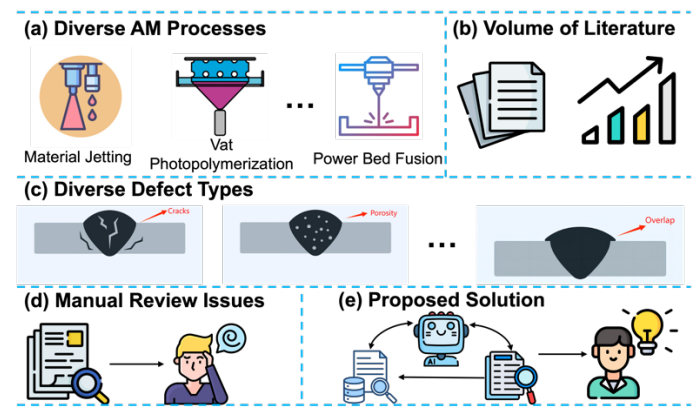

**Figure 1:** (a) Various AM processes, each with unique characteristics. (b) The overwhelming influx of data from AM research. (c) Diverse defects across different AM processes. (d) Challenges of manual review in handling large datasets. (e) The effectiveness of our proposed RAG-based method in addressing these challenges.

Additive manufacturing (AM) has transformed the manufacturing industry by allowing the production of intricate designs, minimizing material waste, and providing exceptional flexibility in design [1, 2] through various manufacturing processes, as depicted in Figure 1a. Despite the progress in AM technologies, the broader adoption of AM faces significant challenges, particularly the presence of defects and anomalies that can undermine the performance and reliability of manufactured parts [3]. Anomalies and defects in AM can be diverse and process-dependent [3] (Figure 1c). For instance, VAT photopolymerization often encounters issues like shrinkage and uneven density [4], Binder Jetting may suffer from slicing and powder spreading defects [5]. For Laser Powder Bed Fusion processes, common defects include porosity, balling, and Surface roughness [6 , 7]. Given the diversity of AM technologies and the range of defects that can occur, identifying and addressing these issues to improve the overall quality and reliability of AM parts is critical.

Generally, anomaly and defect detection have relied on non-destructive testing techniques, and machine learning based methods [3] However, these methods are either labor-intensive, require large amounts of data, are time-consuming, or are often impractical for real-time quality control. Additionally, analyzing test results and conducting quality analysis requires human expertise, which is prone to errors and requires insights from existing scientific papers or company records [8]. The rapid advancement of AM technologies has led to an exponential growth in scientific literature focused on defect and anomaly detection (Figure 1b). While this excess of existing information is a valuable resource, it also creates a significant challenge. These approaches are not only time-consuming but also prone to oversight and error (Figure 1d). Furthermore, interpreting AM processes' anomalies requires domain expertise, making it even more challenging to synthesize findings across diverse research papers [9].

Large Language Models (LLMs) and Multimodal Large Language Models (MLLMs) have been increasingly utilized in additive manufacturing applications [10-13] to enhance process control and anomaly detection [11, 14-16]. For instance, AnomalyGPT [17] is a workflow utilizing MLLM in manufacturing setting, capable of identifying anomalies in manufacturing images through a few-shot inference technique using normal examples. Furthermore, Farimani et al. [11] demonstrated that by employing MLLMs, these models not only can autonomously analyze images of printed layers, identify anomalies such as inconsistent extrusion or layer misalignment, but also adjust printing parameters. This approach aims to enhance the quality of additive manufacturing while reducing the need for human intervention. However, most of these studies manually incorporate expert knowledge or reference images (e.g., images of normal and anomalous parts), which can be prone to oversight and subjectivity. Additionally, they often require fine-tuning on MLLMs, which becomes problematic when dealing with small anomaly datasets [16-18]. Hence, methods such as Contrastive Language-Image Pre-training (CLIP)-based [19] zero-shot models such as AnomalyCLIP [20], WinCLIP [21], M3DM-NR [22], ClipSAM [23], KAnoCLIP [24] have been employed to address this issue. However, their performance is limited when detecting previously unseen manufacturing anomalies, and they rely on predefined anomaly definitions within the CLIP model [25, 26]. Even advanced models like VMAD[27] still suffer from the limitation of requiring manual anomaly definitions, restricting their adaptability to novel anomaly and defect types.

Retrieval-Augmented Generation (RAG) [28] presents a promising approach to addressing these challenges. RAG systems are able to retrieve relevant information from scientific papers, which makes them particularly well-suited for extracting anomaly-related information from extensive AM research [29, 30]. The current research on anomaly detection and classification using retrieved information is limited, primarily focusing on text-based retrieval for anomaly detection [31, 32]. To address this limitation, we propose a novel multimodal RAG-based system specifically tailored for anomaly detection in AM processes. As illustrated in Figure 1e, the system is designed to streamline the extraction of critical information on AM anomalies in both text and image formats [33].Then, it leverages an MLLM to detect and classify anomalies in the test images.

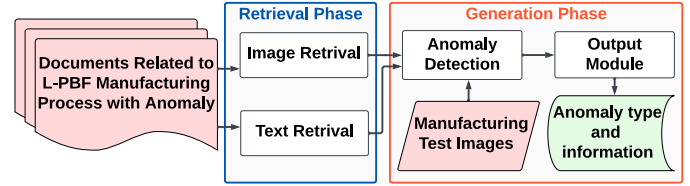

**Figure 2:** Streamlined workflow of the proposed framework for anomaly detection.

The proposed RAG-based framework, as shown in Figure 2, integrates text and image retrieval, classification, and generation models, allowing for the automated extraction and synthesis of information from a wide range of AM research literature, including Vat Photopolymerization, Material Jetting, Binder Jetting, Material Extrusion, Sheet Lamination, Laser Powder Bed Fusion, and Directed Energy Deposition. The presented case studies in this paper specifically focus on Laser Powder Bed Fusion. The contributions of this work are summarized as follows:
1) Proposed a novel multimodal RAG-driven framework (Figure 2) for detecting and classifying anomalies across various AM processes, leveraging both image and text information extracted from scientific papers. This approach addresses key challenges such as literature overload, lack of training data, and the need for manually providing reference images (e.g., similar anomalies or normal images) and information.
2) Developed an end-to-end pipeline that integrates text and image retrieval, classification, and generation models to extract, synthesize, and systematically organize anomaly-related information.

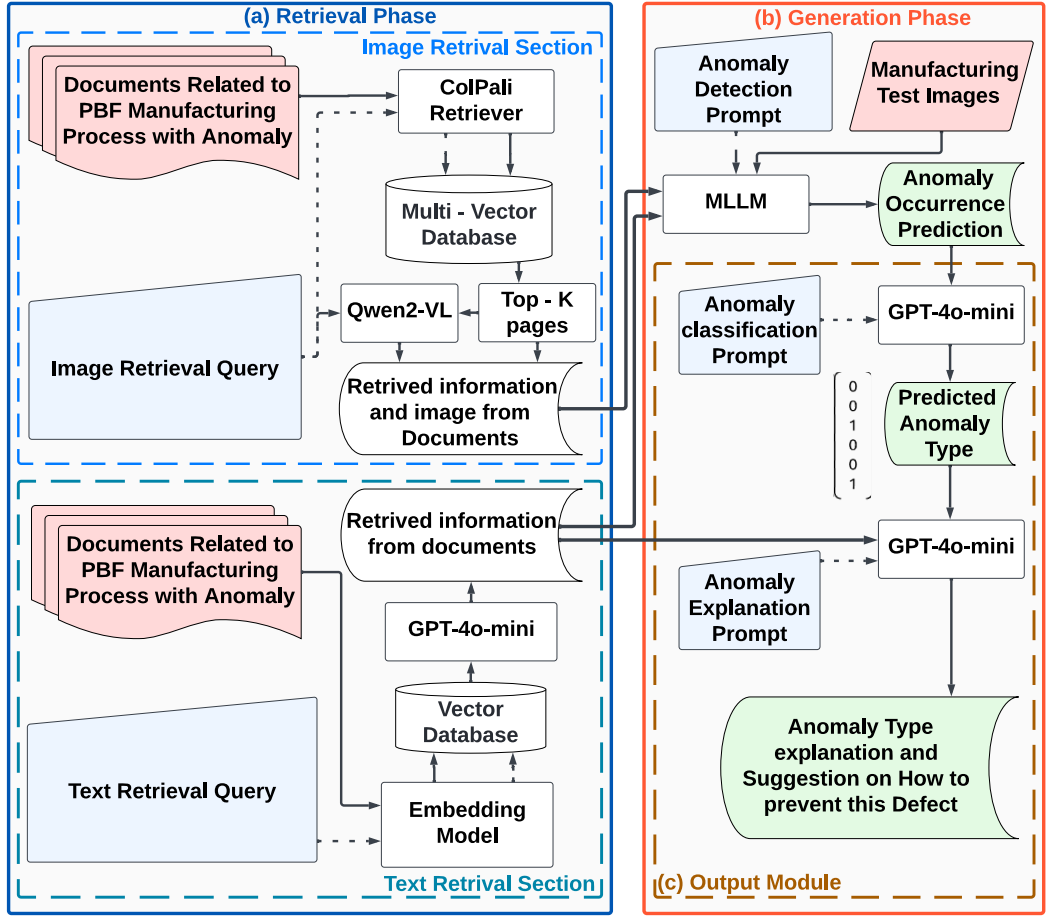

**Figure 3:** Comprehensive workflow of the proposed framework – The framework consists of three main phases: (a) Retrieval Phase, which retrieves relevant images, image descriptions, and textual information related to anomalies detection, root causes, and prevention strategies; (b) Generation Phase, which utilizes the retrieved information to provide the MLLM with sufficient context to detect anomalies, and (c) Output Module, which compiles detected anomalies into a one-hot encoded list of anomalies types, facilitating model accuracy assessment. Additionally, this module provides comprehensive insights, including anomaly occurrence reasons and preventive measures based on detected anomalies in the image.

3) Highlighted the adaptability of the framework, demonstrating its ability to support evolving AM research and emerging processes by applying it to different datasets.
4) Conducted a comparative analysis of the classification capabilities of a small closed-source model and a large open-source model to assess their effectiveness and underlying reasoning in anomaly detection.

The remainder of this paper is organized as follows: Section 2 introduces the proposed methodology. Section 3 presents the dataset used in the case study. Section 4 discusses the results and provides a quantitative assessment of the proposed approach. Section 5 concludes the study.

## 2. METHODOLOGY

The proposed framework consists of two major phases (Figure 3). In the first phase, a dual RAG system is employed to retrieve relevant information about target anomalies. This retrieved data is then passed to the second phase, the generation phase, as a structured prompt. This prompt includes a sample image with the anomaly, a detailed visual description of its characteristics, and contextual information. The prompt will be fed into an MLLM, which predicts the likelihood of the anomalies' presence in the test image. Finally, the output module aggregates all identified anomalies within the image and generates concise yet thorough supplementary information. This ensures a structured and informative summary of anomaly classification. In the following sub-section, each component will be explained in detail.

**2.1 Retrieval Phase**

The retrieval phase utilizes RAG techniques to collect multimodal data essential for anomaly analysis (Figure 3a). This phase consists of two parallel retrieval processes, for visual and textual data, respectively.

**Image Retrieval**: To retrieve related images from documents, ColPali framework [34] is employed to process PDFs containing anomaly-related information, serving as the image retrieval component of the pipeline. Unlike similar methods that depend on optical character recognition, ColPali directly indexes and retrieves visual content from documents [35]. Originally designed to retrieve the most relevant document images for generating text-based responses to queries, this approach can be refined (Figure 4) to specifically retrieve images containing target anomalies, with the *top-k* images saved for further analysis.

Once the relevant images are retrieved, the Qwen2-VL [36] model is employed as the generative component, analyzing the visual data to generate detailed insights or descriptions about the anomaly. Saving the top-ranked image from the retrieval process is essential to enable the image-based detection process in the following generation phase.

**Text Retrieval**: A parallel text-focused RAG pipeline is implemented using GPT-4o-mini [37] and text-embedding-ada-002 [38]. The text-embedding-ada-002 model serves as the embedding component, converting textual information into high-dimensional vector representations that capture semantic meaning and contextual relationships [39]. These embeddings enable efficient similarity searches, allowing the system to identify the most relevant text segments in response to a given query.

This text retrieval process goes beyond the image retrieval section, which primarily focuses on visual characteristics. In addition to analyzing visual aspects of anomalies, it gathers information on detection methods, underlying causes, and prevention strategies. As a result, the framework not only classifies and detects anomalies but also provides a detailed analysis of their origins and potential mitigation measures. Furthermore, if the retrieved image lacks clear anomaly details (e.g., low-resolution images, high-level schematics) or is unavailable, text retrieval provides the model with additional information to accurately detect and classify anomalies. Figure 4 presents the queries used to retrieve necessary information for the following generation phase.

---

**Image Retrieval Query**
1: Retrieve images related to the **{anomaly_name}**, strictly from provided resources. 2: Analyze the retrieved image and include the visual characteristics to help in anomaly identification.

---

**Text Retrieval Query**
Retrieve comprehensive information about **{anomaly_name}**, exclusively from provided resources. Ensure the response includes the following details:
  1. Detailed Description
  2. Common Causes
  3. Visual Characteristics
  4. Prevention Strategies

---

**Figure 4:** Retrieval query for gathering text and image information related to the target anomaly.

---

**Anomaly Detection Prompt**:
Analyze the test image carefully and determine if **{anomaly_name}** is possible. Use the information provided in the reference image and additional scientific information to support your assessment. Provide a clear, short, and reasoned answer with supporting evidence. These are the test images: **{per image: {image_stage_description}: {test_image}}**. The reference image shows an example of **{anomaly_name}**:**{reference_image}**+**{reference_image_description}**. Use it for comparison. Here is additional scientific information about **{anomaly_name}**: **{info_anomaly_text}**.

---

**Anomalies Classification Prompt:**
This is the decision about whether the Anomaly exist: **{detection_results}**. If **{anomaly_name}** is detected in even one of the test images, return 1; otherwise, return 0. Do not provide any additional explanation or reasoning in the response.

---

**Anomalies Explanation Prompt:**
Given the detected anomalies in the manufacturing process: **{classification_results}**, provide a detailed scientific explanation covering the following:
  1. Root Cause
  2. Prevention Strategies
  3. Additional Insights
Ensure the response is precise, technical, and grounded in provided information: **{info_anomaly_text}**

---

**Figure 5:** Generation prompts for anomaly detection and explanation

**2.2 Generation Phase**

In the generation phase (Figure 3b), the multimodal data obtained in the earlier retrieval phase is synthesized and analyzed in detail through both visual recognition and textual analysis. This phase addresses two key objectives: anomaly detection and classification with explanation.

**Anomaly Detection**: Using retrieved anomaly images and textual information, the open-source Qwen2-VL-2B model [40] and the closed-source GPT-4o-mini [37] are utilized to predict anomaly types via the prompt provided in Figure 5. Each MLLM analyzes the visual features of the best retrieved image alongside contextual information from the query to generate an informed classification. These MLLMs are utilized to detect anomalies within dataset images described in Section 3. Given a list of possible anomalies, the detection is performed individually for each anomaly, ensuring the model systematically evaluates each anomaly type based on the provided prompt (Figure 5). The detection process was performed three times per anomaly to provide more robust results for the classification task inside the output model.

**Anomalies Classification and Explanation (Output Module)**: After detecting all anomalies, LLM is employed to synthesize responses for each anomaly and summarizes the results using one-hot encoding classification (prompt provided in Figure 5). The framework then calculates the average prediction for each anomaly type, enhancing the reliability of the evaluation. Next, the system uses the identified anomaly type to guide the subsequent explanation and prevention steps. For this purpose, GPT-4o-mini is integrated as the generative model. It combines the classification results with the text retrieved from the earlier RAG pipeline to synthesize a comprehensive explanation. This explanation includes:
1. A detailed account of the anomaly's characteristics.
2. An analysis of its potential root causes.
3. Preventive measures and recommendations to mitigate similar anomalies in future processes.

This integrated approach ensures anomalies are systematically classified, enabling the system to deliver actionable, context-aware insights tailored to each identified anomaly (Figure 3c).

## 3. MANUFACTURING ANOMALY DATASET

This study focuses on the L-PBF process, given its prominence as a widely used additive manufacturing technique. To evaluate our proposed framework, we utilize an anomaly dataset from Oak Ridge National Laboratory [41] which contains layer-wise powder bed images. Each image may contain one or more of the following anomalies: Recoater Hopping, Recoater Streaking, Incomplete Spreading, Swelling, Debris, Super-Elevation, Soot, Excessive Melting, Localized Bright Spot, Spatter on Powder, Mounding Powder, Localized Dark Regions, or Misprint. These anomalies are annotated in a file within ORNL's dataset. This annotation file is then converted into a text format for each test sample, listing all existing anomalies together, which serves as the ground truth or human reference response. The laser powder-bed fusion (L-PBF) datasets used in this study are sourced from the "EOS M290" and "AddUp FormUp 350" printers, each utilizing different materials, as summarized in Table 1.

Table 1. L-PBF test samples information taken from ORNL's dataset to evaluate the proposed framework [41]

| Printer Make and Model | Material | Test Images |
|---|---|---|
| AddUp FormUp 350 | Maraging Steel | 26 |
| EOS M290 | 17-4 PH Stainless Steel | 14 |
| EOS M290 | DMREF | 9 |
| EOS M290 | Inconel 718 | 5 |

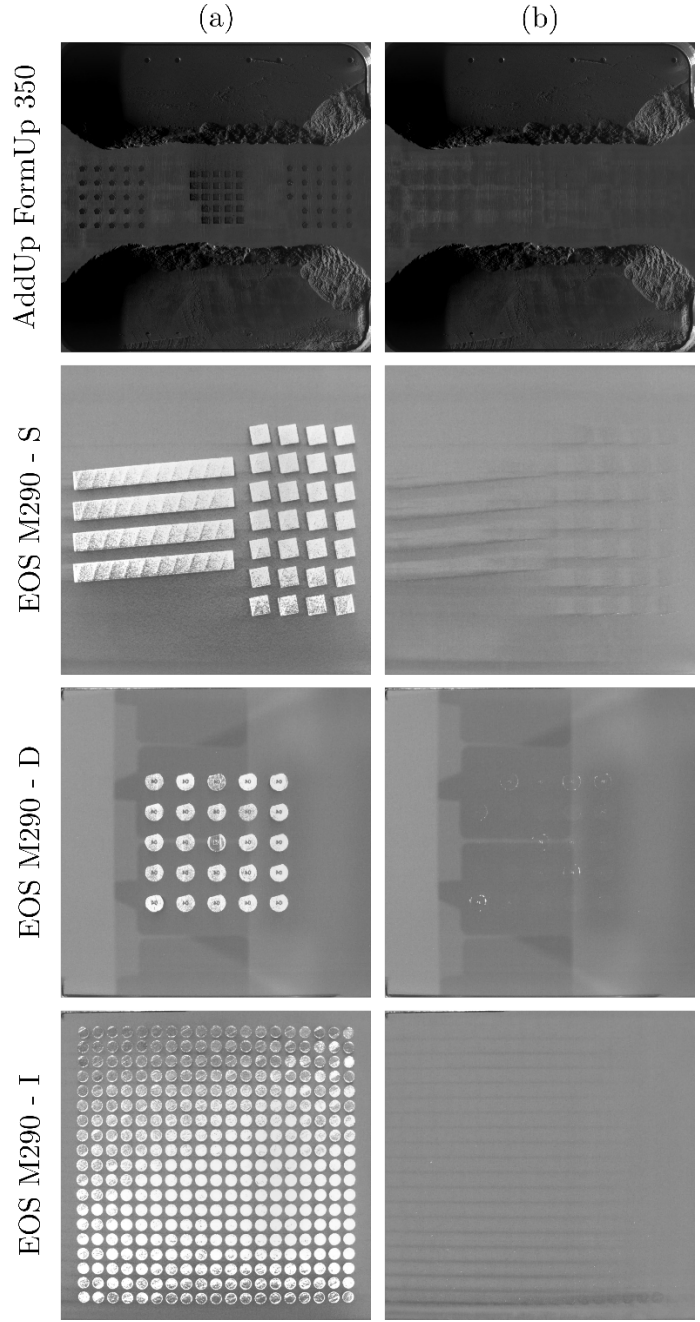

Figure 6: L-PBF's visible light test samples taken from ORNL Dataset [41]: (a) Image captured post-melting (b) Image captured after powder spreading for L-PBF process introduced in Table 1.

These datasets incorporate multi-modal sensor data, including visible light (VL), temporally integrated near-infrared (TI-NR), and wide-band infrared (IR) imaging. Since the dataset includes ground truth files and our proposed model does not require training data, we used these images exclusively for testing. In this study, visible light images were specifically employed for anomaly detection, with one image captured post-melting and the other captured after powder spreading, as shown in Figure 6. For the paper dataset used in our RAG-based study, we focused on articles related to the L-PBF process, a widely adopted additive manufacturing technique. The dataset of document information for RAG (Table 2) consists of scientific papers that investigate various defect types and anomalies within the L-PBF process [6, 42-52]. Table 2 summarizes the anomalies and defects identified in each document, either explicitly stated or conceptually implied, as determined through manual review assisted by an LLM.

Table 2. Dataset of L-PBF papers on defect types and anomalies in the process

| Document | Topic | Available Anomalies and related Defects (Explicitly and Conceptually) | Ref. |
|---|---|---|---|
| Scime, L., et al., *Additive Manufacturing*, 2020. - Oak Ridge National Laboratory (ORNL), 2023. | Layer-wise anomaly detection in PBF | Recoater Hopping, Recoater Streaking, Incomplete Spreading, Debris, Super-Elevation, Spatter on Powder, Jet Misfire, Porosity, Part Damage, Soot Misprint, Localized Dark Regions, Localized Bright Spot, Mounding Powder, Spatter on Powder, Stripe Boundary, Edge Swelling | [6, 41] |
| Sahar, T., et al., *Results in Engineering*, 2023. | ML-based anomaly detection in L-PBF | Porosity, Balling, Cracks, lack of fusion, miscellaneous defect (Recoater Hopping, Part failure), Excessive Melting | [42] |
| Colosimo, B.M. and M. Grasso, *Procedia CIRP*, 2020. | In-situ monitoring in L-PBF: challenges & opportunities | Geometrical Distortions (Misprint, Recoater Hopping, …), Porosity, delamination, Microstructural inhomogeneity, Surface Flaws | [43] |
| Chebil, G., et al., *Journal of Materials Processing Technology*, 2023. | Deep learning for optical monitoring of spatters | Spatter, Lack of fusion, Localized Bright Spot | [44] |
| Peng, X., et al., *Sensors*, 2022. | Multi-sensor fusion for defect detection in PBF | Balling, Porosity, Cracking, Surface Flaws | [45] |
| D'Accardi, E., et al., *Progress in Additive Manufacturing*, 2022. | Detecting and localizing L-PBF defects | Porosity, Surface Flaws, Localized Bright Spot, Lack of fusion | [46] |
| Snow, Z., et al., 2023, Oak Ridge National Laboratory (ORNL). | ML sensor fusion for L-PBF defect detection | Spatter, Excessive Melting, Recoater Streaking, Stripe Boundary, Porosity, Lack-of-Fusion, Localized Bright Spot, Cracks | [47] |
| Cannizzaro, D., et al., *DATE Conference*, 2021. | Image analytics & ML for AM defect detection | Spatter, Incandescence (Excessive Melting, Localized Bright Spot), Horizontal defects (Recoater Streaking), Vertical defects (Recoater Hopping) | [48] |
| Mahmoud, D., et al., *Applied Sciences*, 2021. | ML applications in L-PBF process monitoring | Recoater Hopping, Recoater Streaking, Incomplete Spreading, Debris, Super-Elevation, Spatter on Powder, Overheating (Excessive Melting), Edge Swelling, Curling, Shrinkage, Balling, Under-melting, Porosity, Lack of Fusion, Cracks, Delamination | [49] |
| Mohammadi, M.G. and M. Elbestawi, *Procedia Manufacturing*, 2020. | Real-time monitoring in L-PBF using ML | Porosity, micro-cracks, voids, and Surface Flaws | [50] |
| Okaro, I.A., et al., *Additive Manufacturing*, 2019. | Semi-supervised ML for L-PBF fault detection | Balling, overheating (excessive melting, Localized Bright Spot) | [51] |
| Chicote, B., et al., *Procedia CIRP*, 2022. | Online/offline defect detection in L-PBF | Geometrical Gaps (Misprint), Porosity, Cracks, Lack of Fusion | [52] |

## 4. RESULTS

The dataset introduced in Section 3 is used to evaluate the performance of the proposed method in anomaly detection and classification. Several metrics can be employed to evaluate the accuracy of generated response, including Recall@K, Precision@K, F1 score@K [53-55], Bilingual Evaluation Understudy (BLEU) [56], Recall-Oriented Understudy for Gisting Evaluation (ROUGE) [57], and Embedding-Based Similarity [58], which measures the similarity between the generated and reference text. However, since this study focuses on binary classification (anomaly vs. normal) rather than pure text generation, a clear Yes/No decision is required. Hence, the reference text is converted into a one-hot encoded representation of anomalies (e.g., 1 for anomaly, 0 for normal). The anomaly classification inside the output module then generates the predicted one-hot encoding representation. With this structured format, classification accuracy for each anomaly in the dataset is computed using Equation 1. Also, the overall accuracy for each anomaly is then obtained by averaging the accuracy results across all images.

$$Accuracy = \frac{True\ Positives + True\ Negetives}{Total\ Cases} \quad (1)$$

The framework is tested using Qwen2-VL-2B model and GPT-4o-mini model as the main MLLM for anomaly detection. The accuracy results across all L-PBF's categories, as shown in Figure 7, indicate that GPT-4o-mini outperforms Qwen2-VL-2B model by average margin of 34.6%.

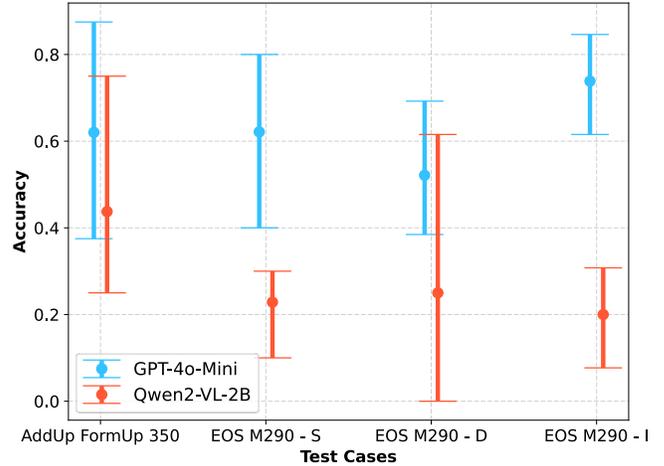

Figure 7: Prediction performance scores across various L-PBF image datasets using different MLLMs.

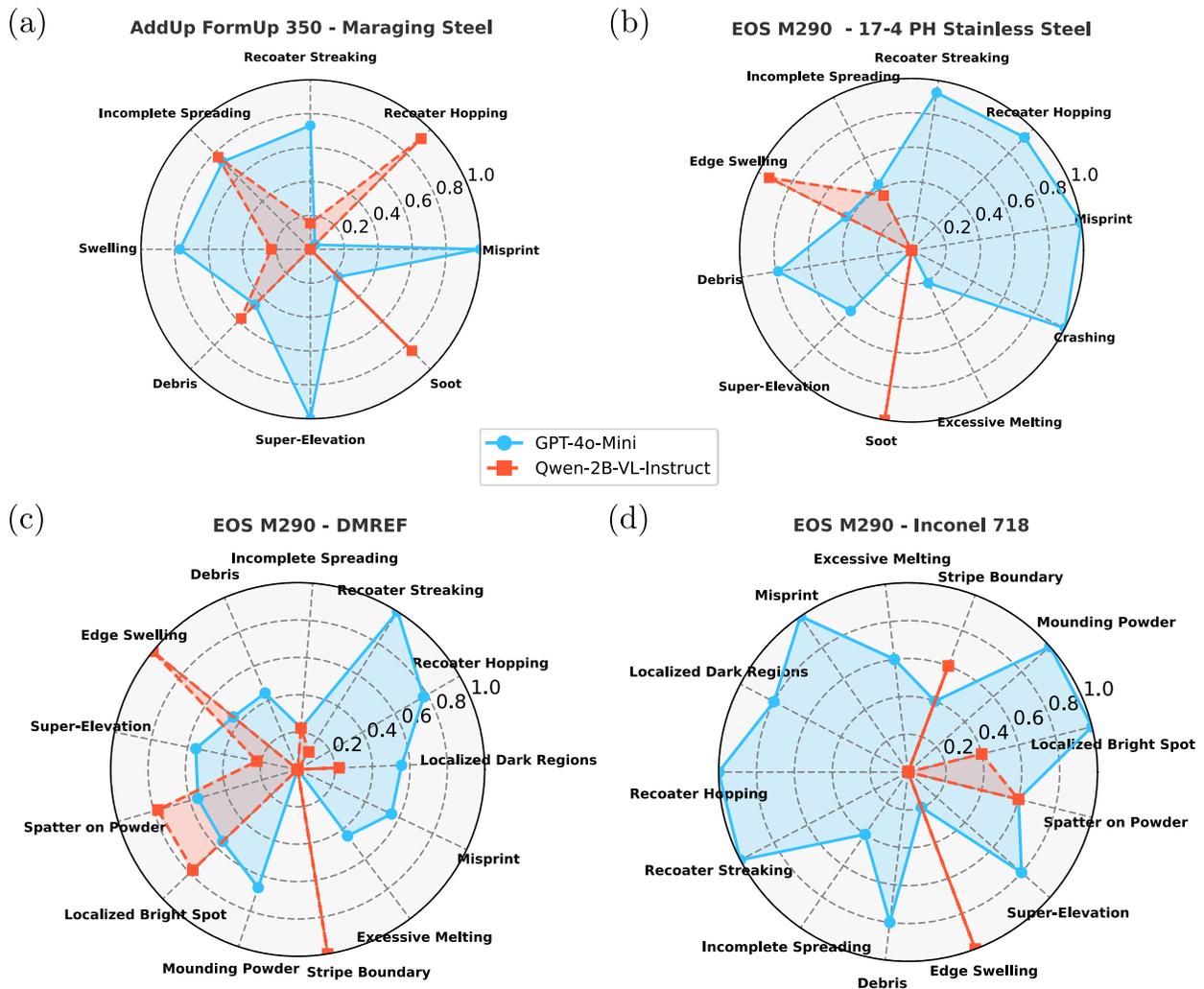

**Figure 8:** The anomaly detection accuracy for desired anomalies for (a) AddUp FromUp 350 (b) EOS M290 - 17-4 PH Stainless Steel (c) EOS M290 – DMREF (d) EOS M290 - Inconel 718

The detailed performance analysis using the same prompt and input images, as shown in Figure 8 and Tables A1–A4 in the Appendix, indicates that Qwen2-VL-2B does not outperform the proportional random baseline and produces mostly the same results. This suggests that, given a detection prompt, Qwen2-VL-2B consistently predicts that an anomaly exists, even in cases where no anomaly is present. In other words, it tends to flag every instance as anomalous rather than differentiating between normal and anomalous cases. Furthermore, the proportional random baseline [59] reveals that anomalies are not evenly distributed within the dataset. This imbalance is expected, as the instrument in each ORNL's dataset [41] remains the same, causing some anomalies to appear frequently while others are rarely present. This dataset imbalance underscores the importance of using the accuracy metric defined in Equation 1, which accounts for cases where the MLLM correctly detects the absence of anomalies. Without this consideration, a model like Qwen2-VL-2B, which naively predicts the presence of anomalies, would achieve a high F1-score and sensitivity, misleadingly inflating its performance evaluation.

On the other hand, GPT-4o-mini, using the same input prompt, demonstrates a significantly better ability to differentiate between anomalous and normal cases. As a result, it consistently achieves higher average accuracy compared to both the proportional random baseline [59] and the Qwen2-VL-2B model. However, GPT-4o-mini still encounters challenges in certain cases, such as detecting edge swelling, strip boundaries, and soot. These challenges may arise from inadequate retrieved information for the detection task or the MLLM's difficulty in identifying specific anomalies. Due to the black-box nature of LLMs [60], pinpointing the exact cause remains challenging. A key advantage of GPT-4o-mini is that when it fails to detect an anomaly, it does not overdetect like Qwen2-VL-2B. Instead, it tends to avoid incorrect anomaly detection, ultimately leading to a more reliable accuracy assessment.

Upon detecting and classifying the anomalies in all images, the framework generates an explanation detailing why the anomalies occur, followed by preventive measures to mitigate their impact. Figure 9 showcases an example from the EOS M290 – 17-4 PH stainless Steel dataset, including the ground truth, predicted anomalies, and explanations for the predicted anomalies.

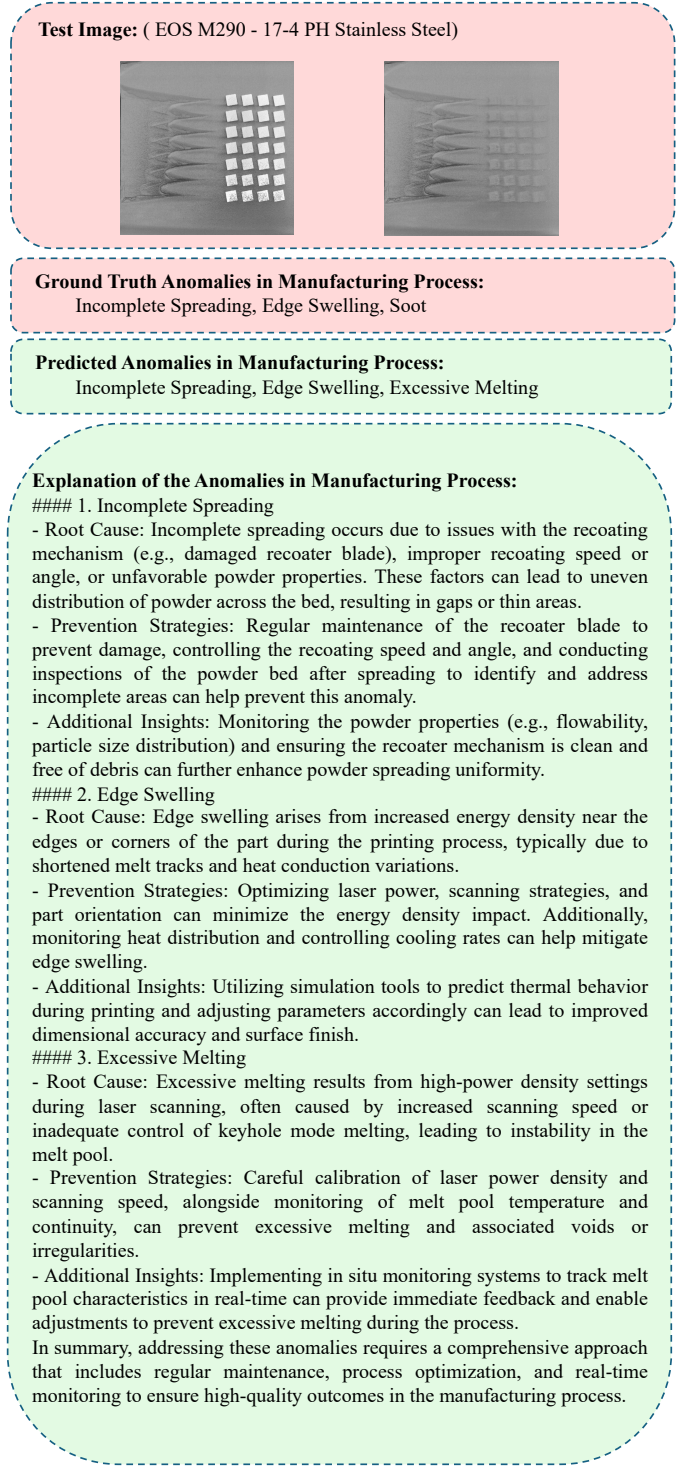

Figure 9: Predicted anomalies and explanation given test image from EOS M290 – 17-4 PH stainless Steel

Sometimes, adding an RAG system can introduce more problems than it solves regarding hallucination [61]. The RAG system design, which bases outputs on retrieved information from other sources to increase accuracy, can, in some cases, exacerbate the issue [62]. If a RAG component is improperly configured [63, 64], it can reduce reliability by creating a false sense of accuracy. To ensure this issue did not occur in this study, an additional evaluation was conducted on GPT-4o-mini, the best-performing MLLM, to assess the effects of incorporating a RAG system for additional information retrieval. The results in Table 2 show that incorporating RAG systems enhances prediction accuracy compared to models relying solely on their pre-trained knowledge. This finding aligns with existing research, which also demonstrates improved performance when scientific literature is leveraged.

Table 3. Comparison of average accuracy with and without the retrieved information

| Test Case Dataset | With Retrieval | Without Retrieval |
|---|---|---|
| AddUp FromUp 350 | **0.620** | 0.610 |
| EOS M290 - S | **0.621** | 0.471 |
| EOS M290 – D | **0.521** | 0.401 |
| EOS M290 - I | **0.738** | 0.523 |

## 5. CONCLUSION

The proposed RAG-based framework offers a novel and efficient solution to the challenge of anomaly detection in AM. By integrating advanced multimodal retrieval and generation models, our system enables automated and context-aware identification and classification of anomalies in AM processes. By combining image and text retrieval with state-of-the-art generative models, we demonstrate the feasibility of establishing an anomaly detection and classification model solely based on literature-based information, without requiring in-house experimental data. Additionally, the framework's ability to continuously integrate new research literature ensures its adaptability to evolving AM technologies.

Future work will focus on improving the model's accuracy by incorporating more advanced image retrieval models and refining the multimodal generation process to mitigate the black-box nature of anomaly and defect detection and enhance the accuracy. Despite the proposed model's current limitations in classification and detection accuracy, this framework shows promise as a robust and practical tool for real-time defect detection in additive manufacturing. It holds significant potential to advance automated quality control in the industry.

## ACKNOWLEDGEMENT

The authors acknowledge financial support from the National Science Foundation grants CMMI-2414398, CMMI-2001081, CMMI-2336448, CMMI-2434519, and DMR-2102406. KNK also gratefully acknowledges the Pratt & Whitney Institute for Advanced Systems Engineering Fellowship from the University of Connecticut. Additionally, the authors acknowledge the Oak Ridge Leadership Computing Facility (OLCF) at Oak Ridge National Laboratory, managed by UT-Battelle, LLC for the U.S. Department of Energy under contract DE-AC05-00OR22725, for providing access to the dataset utilized in this study.

## APPENDIX

Table A1. Performance Metric for AddUp FromUp 350

| Anomaly | Random Baseline | Accuracy | |
|---|---|---|---|
| | | Qwen2-VL-2B | GPT-4o-Mini |
| Recoater Hopping | 0.96 | 0.92 | 0.19 |
| Recoater Streaking | 0.15 | 0.15 | 0.58 |
| Incomplete Spreading | 0.77 | 0.77 | 0.73 |
| Swelling | 0.23 | 0.23 | 0.81 |
| Debris | 0.58 | 0.58 | 0.5 |
| Super-Elevation | 0 | 0 | 0.96 |
| Soot | 0.85 | 0.85 | 0.23 |
| Misprint | 0 | 0 | 0.88 |

Table A2. Performance Metric for EOS M290 - S

| Anomaly | Random Baseline | Accuracy | |
|---|---|---|---|
| | | Qwen2-VL-2B | GPT-4o-Mini |
| Recoater Hopping | 0 | 0 | 0.93 |
| Recoater Streaking | 0 | 0 | 0.93 |
| Incomplete Spreading | 0.36 | 0.36 | 0.43 |
| Edge Swelling | 0.93 | 0.93 | 0.43 |
| Debris | 0 | 0 | 0.79 |
| Super-Elevation | 0 | 0 | 0.5 |
| Soot | 1 | 1 | 0 |
| Excessive Melting | 0 | 0 | 0.21 |
| Crashing | 0 | 0 | 1 |
| Misprint | 0 | 0 | 1 |

Table A3. Performance Metric for EOS M290 - D

| Anomaly | Random Baseline | Accuracy | |
|---|---|---|---|
| | | Qwen2-VL-2B | GPT-4o-Mini |
| Recoater Hopping | 0 | 0 | 0.78 |

| Anomaly | Random Baseline | Qwen2-VL-2B | GPT-4o-Mini |
|---|---|---|---|
| Recoater Streaking | 0.11 | 0.11 | 1 |
| Incomplete Spreading | 0.22 | 0.22 | 0.22 |
| Debris | 0 | 0 | 0.44 |
| Edge Swelling | 1 | 1 | 0.44 |
| Super-Elevation | 0.22 | 0.22 | 0.56 |
| Spatter on Powder | 0.89 | 0.78 | 0.56 |
| Localized Bright Spot | 0.78 | 0.78 | 0.56 |
| Mounding Powder | 0 | 0 | 0.67 |
| Stripe Boundary | 1 | 1 | 0 |
| Excessive Melting | 0 | 0 | 0.44 |
| Misprint | 0 | 0 | 0.56 |
| Localized Dark Regions | 0.22 | 0.22 | 0.56 |

Table A4. Performance Metric for EOS M290 – I

| Anomaly | Random Baseline | Accuracy | |
| | | Qwen2-VL-2B | GPT-4o-Mini |
|---|---|---|---|
| Recoater Hopping | 0 | 0 | 1 |
| Recoater Streaking | 0 | 0 | 1 |
| Incomplete Spreading | 0 | 0 | 0.4 |
| Debris | 0 | 0 | 0.8 |
| Edge Swelling | 1 | 1 | 0.2 |
| Super-Elevation | 0 | 0 | 0.8 |
| Spatter on Powder | 0.6 | 0.6 | 0.6 |
| Localized Bright Spot | 0.6 | 0.4 | 1 |
| Mounding Powder | 0 | 0 | 1 |
| Stripe Boundary | 0.6 | 0.6 | 0.4 |
| Excessive Melting | 0 | 0 | 0.6 |
| Misprint | 0 | 0 | 1 |
| Localized Dark Regions | 0 | 0 | 0.8 |